\pdfoutput=1

\documentclass[11pt]{article}

\usepackage{acl}

\usepackage{times}
\usepackage{latexsym}
\usepackage{subcaption}
\usepackage{algpseudocode}
\usepackage{listings}
\usepackage{amsmath}
\usepackage[T1]{fontenc}

\usepackage[utf8]{inputenc}

\usepackage{microtype}

\usepackage{inconsolata}

\usepackage{times}
\usepackage{latexsym,amssymb}
\usepackage{booktabs}
\usepackage{amsmath}
\usepackage{comment}
\usepackage{graphicx}
\usepackage{stmaryrd}
\usepackage{bm}
\usepackage{dsfont}
\usepackage{xspace}
\usepackage{longtable}
\usepackage{algorithm2e}
\usepackage{multirow,multicol}
\usepackage{lscape}
\usepackage{tabularx}
\usepackage[capitalize]{cleveref}
	
\usepackage{soul}
\usepackage{pdflscape}
\usepackage{afterpage}

\usepackage[disable]{todonotes}
\newcommand{\ourmethod}{\textsc{OPEN}\xspace}
\newcommand{\bzl}[2][]{\todo[color=red!25, #1]{\textbf{bzl}: #2}}

\newcommand{\kh}[2][]{\todo[color=gray!25, #1]{\textbf{kh}: #2}}

\title{Bayesian Preference Elicitation with Language Models}

\author{
  Kunal Handa \\
  University of Oxford \\
  \texttt{kunal.handa@cs.ox.ac.uk} \And
  Yarin Gal \\
  University of Oxford \\
  \texttt{yarin@cs.ox.ac.uk} \And
  Ellie Pavlick \\
  Brown University \\
  \texttt{ellie\_pavlick@brown.edu} \AND
  Noah Goodman \\
  Stanford University \\
  \texttt{ngoodman@stanford.edu} \And
  Jacob Andreas \\
  MIT \\
  \texttt{jda@mit.edu} \And
  Alex Tamkin\thanks{Collaborative advising; co-last authors.} \\
  Anthropic \\
  \texttt{atamkin@anthropic.com} \And
  Belinda Z. Li$^*$ \\
  MIT \\
  \texttt{bzl@mit.edu} \\
  }

\begin{document}
\maketitle
\begin{abstract}

Aligning AI systems to users' interests requires understanding and incorporating humans' complex values and preferences. 
Recently, language models (LMs) %
have been used to \emph{gather information} about the preferences of human users. This preference data can be used to fine-tune or guide other LMs and/or AI systems.
However, LMs have been shown to struggle with crucial aspects of preference learning: quantifying uncertainty, modeling human mental states, and asking informative questions. These challenges have been addressed in other areas of machine learning, such as Bayesian Optimal Experimental Design (BOED), which focus on designing informative queries within a well-defined feature space. But these methods, in turn, %
are difficult to scale and apply to real-world problems %
where simply identifying the relevant features can be difficult. We introduce \textbf{\ourmethod} (\textbf{O}ptimal \textbf{P}reference \textbf{E}licitation with \textbf{N}atural language) a framework that uses BOED to guide the choice of informative questions and an LM to extract features and translate abstract BOED queries into natural language questions. By combining the flexibility of LMs with the %
rigor of BOED, \ourmethod can optimize the informativity of queries while remaining adaptable to real-world domains. In user studies, we find that \ourmethod outperforms existing LM- and BOED-based methods for preference elicitation.

\end{abstract}

\section{Introduction}

\begin{figure*}
    \centering
    \includegraphics[width=\linewidth,trim={0 11cm 0 0},clip]{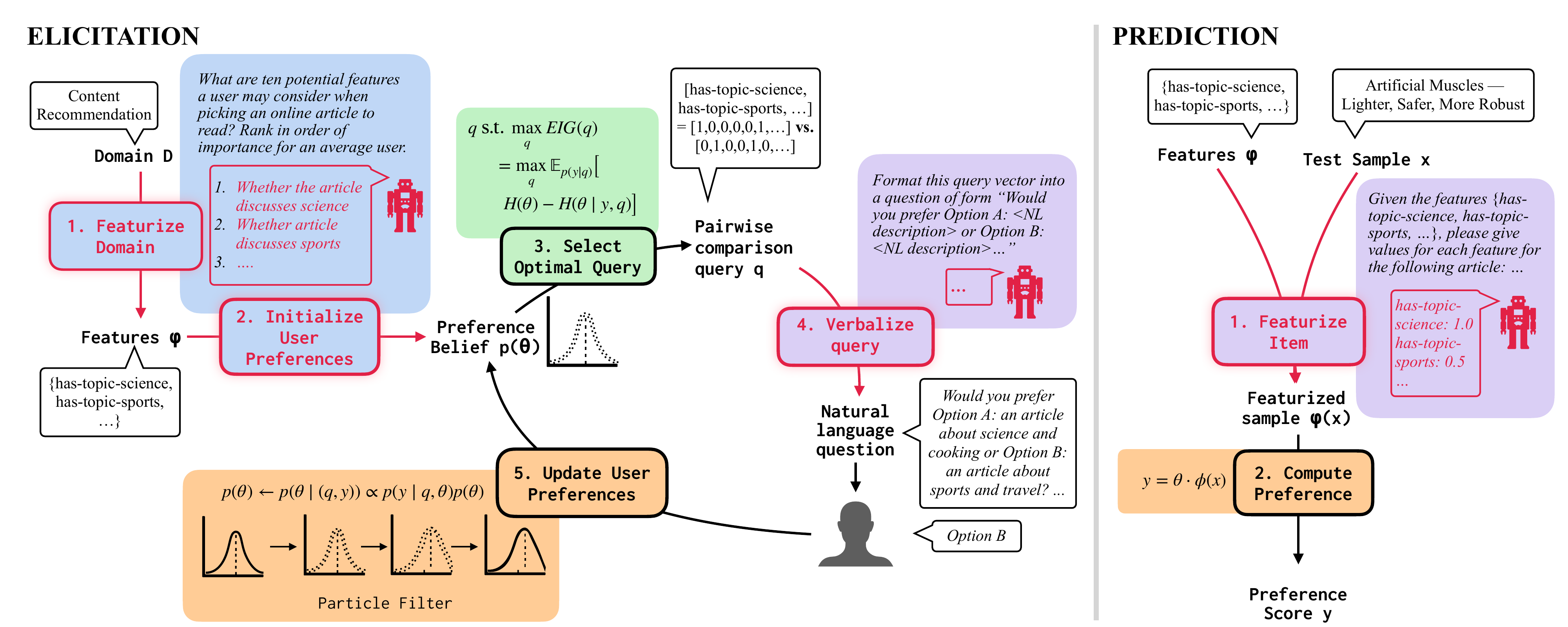}
    \caption{Overview of the \ourmethod framework. In \textcolor{red}{red} are the parts where we use a language model. During the \textit{elicitation} stage, first, a domain $D$ is featurized into feature $\phi$ with a language model, which also gives us a ranking of importance over features (which is used to initialized a prior $p(\theta)$ over user preferences). Based on the prior user preferences, the optimal pairwise comparison query $q$ is computed, which is then verbalized using an LM into natural language. The user response is then taken to update the prior over beliefs. During the \textit{prediction} stage, a LM is used to featurize a test sample according to the featurization $\phi$ derived from the elicitation stage, then a preference score is computed using the elicited preferences $\theta$.}
    \label{fig:framework}
\end{figure*}

Understanding users' complex preferences and requirements is necessary for %
accurately and safely automating real-world tasks. 
Modeling preferences in a domain of interest requires both understanding \emph{what features} of the domain are relevant to model, and \emph{how important} these features are relative to each other \citep{lin2022inferring, lindner2022interactively, Sadigh2017ActivePL, Fürnkranz2012}.
For example, when building a content recommendation system, relevant features may be different article topics---e.g.\ science, politics, or celebrity culture.
A user may decide whether or not to read an article based on the strength of their preferences along these different axes.

With the widespread use of preference data to align language models (LMs) with human users, there has been growing interest in using LMs themselves to elicit information about human preferences.
Past work focused on
prompting LMs to ask questions about user preferences~\citep{li2023eliciting,lin2023decision, piriyakulkij2023active} has shown that LMs are capable of identifying decision-relevant features in domains spanning content recommendation, software engineering, and moral reasoning.
Nonetheless, prompting offers limited control over LMs' information-gathering strategies and often fails to produce questions that are useful or informative.

By contrast, a long line of work on optimal experimental design, such as dueling bandits~\cite{10.1214/18-AOS1772} and Bayesian preference learning approaches~\cite{foster2019variational, gal2017deep, houlsby2011bayesian}, has developed methods to efficiently infer user preferences from a limited number of interactions.
But these methods also have their limitations---they have primarily been applied to tasks with simple, highly-structured data where feature spaces are constrained and interactions are short. How can we get the best of both worlds? \textit{Can we learn similarly sample-efficient models of  user preferences in complex, open-ended tasks?}

In this paper, we introduce \textbf{O}ptimal \textbf{P}reference \textbf{E}licitation with \textbf{N}atural language \textbf{(\ourmethod)}---a framework that 
leverages the complementary advantages of LMs and Bayesian Optimal Experiment Design (BOED) methods (see~\Cref{fig:framework}).

\ourmethod uses an LM to select the relevant features the user likely cares about (\Cref{fig:framework}, \textit{steps 1, 2}), and a Bayesian model %
to select optimal pairwise comparison queries (\textit{step 3}), which are translated into natural language by a LM (\textit{step 4}). In \ourmethod, we utilize a LM to provide feature coverage and a natural conversational interface with the  user (\textit{steps 1, 2, 4}), while leveraging a Bayesian model to track feature weightings and select informative questions (\textit{steps 3, 5}).
Using \ourmethod to elicit user preferences in a content recommendation domain, we find that it outperforms both LM- and BOED-based preference elicitation approaches.

\section{Preliminaries \& Background}
\label{sec:task}
We aim to actively model human preferences through iterative interactions with a user. Over the course of a conversation, we construct and update a model through a series of user queries, each designed such that they (1) \textbf{maximize information gain} of the users' preferences and (2) can reliably be answered by users and \textbf{elicit informative responses}. 

\subsection{Bayesian Optimal Experimental Design}
We use Bayesian Optimal Experimental Design (BOED; \citealp{lindleyinformation, lindleybayesian, rainforth2023modern}) a mathematical abstraction for selecting optimal queries in an active learning setup, as the basis for our interaction framework. BOED is a common approach to preference learning and function estimation \cite{foster2019variational, gal2017deep, houlsby2011bayesian} and has been used across many different domains and disciplines \cite{cavagnaro2009, DEHIDENIYA2018277, Dushenko2020, Vanlier2012}.

In BOED, the goal is to select an experimental design that maximizes the information gain for parameters of interest. We begin with a predictive model $p(y \mid q, \theta)$ which defines the relationship between an experiment 
$q$, an experimental outcome 
$y$, and the parameters of interest 
$\theta$. The prior $p(\theta)$ describes an initial belief about the parameters.\bzl{in our setting $p(\theta)$ isn't just prior, but belief about $\theta$ before the current turn.}

We can formalize the information we gain about $\theta$ from each experiment as: 
\begin{equation}
   \text{IG}(q,y) = H[\theta] - H[\theta\mid y, q],
\end{equation}
where $H$ is the Shannon entropy.
However, because this notion of information gain relies on $y$, the outcome of the experiment, we cannot select the optimal experiment until after the outcome has been observed. To find the optimal design, we instead take the expectation of the information gain for the marginal distribution of $y$ over all possible outcomes $p(y \mid q) = \mathbb{E}_{p (\theta)}[p(y \mid \theta, q)]$,\footnote{ This is equivalent to the mutual information between $\theta$ and $y$ given $q$. } yielding: 
\begin{equation}
    \label{eqn:EIG}
    \text{EIG}(q) = \mathbb{E}_{p(y\mid q)}[\text{IG}(q, y)].
\end{equation}
    
Thus, the optimal experiment can be defined as:
\begin{equation}
    \label{eqn:optimal-EIG}
    q^* = \text{argmax}_q(\text{EIG}(q)).
\end{equation}

In our preference learning setting, we want to pick the pairwise comparison question ($q$) that will allow us to learn about the user's preferences ($\theta$) as efficiently as possible.
At each interaction, $y$ is the user's response to $q$. BOED is uniquely powerful in our sequential setting---enabling us to, after each interaction, incorporate users' answers ad-hoc to greedily update our belief of their preferences.

\subsection{Modeling Human Preferences}
\label{sec:modeling-prefs}
Following previous preference learning literature \cite{lin2022inferring, lindner2022interactively, Sadigh2017ActivePL, Candeal-Haro1995}, we model human preferences about examples $x$ as linear over the features of $x$.
We suppose there is a feature function $\phi$ that maps examples $x$ into a feature space $\mathbb{R}^d$, over which the user has a set of preference weights $\theta \in \mathbb{R}^d$. User preferences can then be modeled in terms of $\theta\cdot \phi(x)$. Specifically, we use a Bradley-Terry model \cite{BradleyTerry} of human pairwise preferences, where given a choice between $x_a$ and $x_b$, the probability the user chooses one over the other is given by: 
\begin{align}
\label{eq:bradley_terry}
\begin{split}
    p(x_a&\mid (x_a,x_b), \theta) \\&=\sigma(\theta\cdot\phi(x_a) - \theta\cdot\phi(x_b)) \\
    p(x_b&\mid (x_a,x_b), \theta) \\&=\sigma(\theta\cdot\phi(x_b) - \theta\cdot\phi(x_a))
\end{split}
\end{align}

\section{The \ourmethod Framework}
\label{sec:framework}

Existing Bayesian and bandit-based preference learning methods assume that $\phi$ is known \emph{a priori}, and that examples can be straightforwardly featurized.
How should we apply these methods in real-world domains like content recommendation where features themselves are difficult to extract and translate into concrete instances? %
To operationalize BOED and other uncertainty-based preference learning approaches in complex, real-world domains, we introduce \ourmethod: a domain-agnostic preference learning framework that combines LMs with BOED. \ourmethod %
uses LMs to identify environment-specific features and %
interface with a user, while using BOED to maintain principled approach to question-selection and decision-making.

At a high level,
\ourmethod learns user preferences by following the below steps:
\begin{enumerate}
    \item \textbf{Featurization}: Using a description of the domain, $D$, the LM extracts a set of NL features which we use to define the feature function $\phi$. 
    \item \textbf{Initializing User Preferences}: We prompt the LM to rank the NL features and set a prior over user preferences, $p(\theta)$, according to this ranking.
    \item \textbf{Selecting an Optimal Question}: The Bayesian model samples all possible pairwise comparison questions and selects the maximally informative question, $q^*$ 
    using \Cref{eqn:optimal-EIG}.
    \item \textbf{Verbalize query}: The LM translates $q^*$ into a NL question for the user (this can also be thought of as $\phi^{-1}$).
    \item \textbf{Update User Preferences}: Given the user's response to the question, we compute the Bayesian posterior $p(\theta \mid y, q)$.
    \item \textbf{Prediction}: The Bayesian model predicts the user's response to each test case, using the Bradley--Terry model of human preferences. 
\end{enumerate}

Even for the linear preference model described in \cref{sec:modeling-prefs}, computation of the posterior
$p(\theta \mid y, q)$ (and thus 
computation of EIG($q$)) is intractable.
We implement a tractable approximation using a Bayesian Particle Filter \cite{Gordon_1993, Doucet2008ATO, elfringparticle}. As each particle can be thought of as a plausible instantiation of user preferences $\theta$, we refer to individual particles as \textbf{personas} $\mathbf{p}_i$.

Crucially, though, \ourmethod does not necessitate the use of particle filtering or even BOED. Our framework fundamentally provides a domain-agnostic approach to active and principled preference learning: enabling any uncertainty-based preference-learning algorithm %
to interface with a user in NL.

An illustration of our approach can be found in~\Cref{fig:framework}.
Below, we detail our implementation of the aformentioned components of \ourmethod. 

\subsection{Featurization}
\label{sec:featurization}
We prompt the LM with a general description of the domain of interest (e.g. content recommendation of news articles), and ask it to produce natural language descriptions of pertinent features in that domain, $\mathcal{F}$. For example, in the content recommendation domain, the LM outputs ``\textit{whether the article discusses science}'' as a pertinent feature. This list of natural language features can then be converted to a function $\phi$ that transforms test samples to feature values by prompting another LM with $\mathcal{F}$ and a test sample $x$ (see~\Cref{sec:methods-prediction}).

\subsection{Initializing User Preferences}
\label{sec:initialization}
When prompting the LM for features in 
\Cref{sec:featurization}, we simultaneously ask it to \textit{rank} features from \textit{most} to \textit{least} important.
These rankings are then used to
initialize the prior for \ourmethod's belief of the user's preference, $p(\theta)$. At a high level, we assume that more important features are also more impactful for humans' decision making about their preferences, and thus, we should model greater diversity over them compared to less important features.
Because we assume that $\theta$ is linear in the feature space, we model $p(\theta)$ as a Bayesian linear model\bzl{did we talk about this anywhere until now?} with each feature weight parameterized by a normal distribution:
\begin{align}
\begin{split}
    p(\theta_i) \sim \mathcal{N}(0, \sigma^2) \cdot w_i, \\
    w_i = 1.2 - 0.12i
\end{split}
\end{align}
where $\sigma^2$ is the base variance for each feature before scaling, and $w_i$ captures the relative importance of features from highest to lowest.

We construct our initial sample of $N$ personas by sampling each element of each persona independently from $p(\theta_i)$.

\subsection{Selecting the Optimal Question}
\label{sec:question-selection}
To efficiently learn user preferences, we aim to select the pair of options that maximize the EIG with regards to the user's preferences at each interaction step. This process is grounded in the BOED framework, as described earlier. 
Given a set of features, $\mathcal{F}$, we define the space of all possible pairwise comparison questions as: 
\[
    \begin{aligned}
    \mathcal{Q} = \{(\mathbf{o}_a, \mathbf{o}_b) | \mathbf{o}_a, \mathbf{o}_b \in \{0,1\}^{|\mathcal{F}|},\\
    \sum_{i=1}^{|\mathcal{F}|} (\mathbf{o}_a)_i = \sum_{j=1}^{|\mathcal{F}|} (\mathbf{o}_b)_j = K\},
    \end{aligned}
\]
where each option $\mathbf{o}_a$ and $\mathbf{o}_b$ is represented as a binary vector indicating the presence (1) or absence (0) of each feature in the comparison. In our experiments, we set total number of features $|\mathcal{F}|=10$ and the number of features compared per query $K=2$.\footnote{From preliminary testing, we found that including ten total features ($|\mathcal{F}|=10$) and two features ($K=2$) in each option best balanced users' mental effort with the informativity of the question.} The goal is to select the question $(\mathbf{o}_a^*, \mathbf{o}_b^*) \in \mathcal{Q}$ that maximizes the EIG about the user's preferences $\theta$.

Recalling Equation \ref{eqn:optimal-EIG}, the EIG for the optimal pairwise comparison $(\mathbf{o}_a^*, \mathbf{o}_b^*)$ with which to query the user is then given by:
\[
    (\mathbf{o}_a^*, \mathbf{o}_b^*) = \text{argmax}_{(\mathbf{o}_a, \mathbf{o}_b) \in \mathcal{Q}} \text{EIG}(\mathbf{o}_a, \mathbf{o}_b)
\]

\subsection{Querying the User: Mapping Pairwise Comparisons to NL}
\label{sec:mapping}
Given the optimal pairwise comparison $(\textbf{o}_a, \textbf{o}_b)$ and natural language feature descriptions $\mathcal{F}$, we prompt a LM to convert the feature vectors $\textbf{o}_a, \textbf{o}_b$ into a NL question for the user.
We also prompt the LM to \textit{synthesize two examples} in accordance with those feature vectors, which is equivalent to taken the inverse of the featurization function $\phi^{-1}$.
We display these examples in the user interface alongside each option in the pairwise comparison question. 

\subsection{Posterior Update}
\label{sec:update}

At each interaction step $t$, having observed the user's selection $y_t$ between $(\mathbf{o}_a, \mathbf{o}_b)$, we update our belief about the user's preferences ($\theta$) based on their response ($y_t$) to the pairwise comparison question ($q_t$) by computing the posterior:
\begin{align}
\label{eq:update}
& p(\theta \mid y_t, q_t, \mathcal{H}) = \frac{p(y_t \mid q_t, \theta) \cdot p(\theta\mid \mathcal{H})}{p(y_t, q_t, \mathcal{H})},
\end{align}
where $\mathcal{H} = \{(q, y)\}_{0\cdots t-1}$ denotes the conversation history (prior sequence of questions and answers).
We approximate the update in ~\Cref{eq:update} using our particle filter. At each step $t$, we reweight each particle $\mathbf{p}_i$ by 
\begin{align}
\begin{split}
    w_i &= p(y_t| \mathbf{o}_a, \mathbf{o}_b, \mathbf{p}_i^\top) \\
    & = \sigma(\mathbf{p}_i^\top \phi(\mathbf{o}_a) - \mathbf{p}_i^\top \phi(\mathbf{o}_b)).
\end{split}
\end{align}
Then, we fit a Gaussian to the distribution of $w_i \mathbf{p}_i$s. Finally, we resample new personas from this distribution.
For a detailed overview of our algorithm, see~\Cref{sec:app-particle_filter}.

\subsection{Prediction}
\label{sec:methods-prediction}
After each user interaction, we evaluate our model via a set of test cases, which are also presented to the user at the end of the study. 
For our test cases, we use pairwise comparison questions $(x_a,x_b)$, where each $x_a, x_b$ are taken from real-world examples from the domain (e.g. for content recommendation, the headline and lede of a New York Times news article).
For each test-case question, we select $y\in [x_a,x_b]$ according to:
\begin{align*}
\max_y \mathbb{E}_{\theta} \left[p(y\mid\theta, (x_a,x_b))\right]
\end{align*}
(where the expectation is taken over our distribution of $p(\theta)$ at that point in time).
Recall preference probabilities $p(y\mid\theta, (x_a,x_b))$ are defined following the Bradley-Terry model (\Cref{eq:bradley_terry}).
To compute $\phi(x_a)$ and $\phi(x_b)$ (recall we were given a set of natural language descriptions of features $\mathcal{F}$), we prompt a LM to give us concrete values for each features, based on the natural language feature descriptions and the test-set sample.

\section{Experimental Setup}

We use \ourmethod to investigate how incorporating LMs alongside explicit information-theoretic computations (via BOED) can improve existing preference-learning methodology and provide a novel NL interface with which to approach personalized machine learning. 

We outline \ourmethod's experimental details, the three different baselines which we compare to, the evaluation framework, and the human participants' interface. 

\subsection{Hyperparameters}
In the featurization step, we query the LM for $|\mathcal{F}|=10$ features to define the domain.
When sampling pairwise comparison questions, each option contains $K=2$ features. 
We use GPT-4 as the LM for all of our experiments (see \Cref{app:prompts} for additional details).

\subsection{Baselines} 
\label{sec:baselines}
We consider three baselines: 

\paragraph{LM-only Open-Ended Questions} Following \citet{li2023eliciting}'s best-performing method in their content-recommendation setting, we prompt a LM to ask informative, open-ended questions, without any explicit optimization. The LM is provided the complete conversation history of the interaction before asking each subsequent question. The user is able to answer the question in free-form natural language.

\paragraph{LM-only Pairwise Comparison Questions} We prompt a LM to ask an informative, pairwise comparison question, without any explicit optimization. The LM is provided the complete conversation history of the interaction and the feature set from Section \ref{sec:featurization} before asking each subsequent question. To answer the question, the user selects either `Option A' or `Option B' based on their preference.

\paragraph{User Self-Mapping Pairwise Comparisons to NL} We provide the user the feature vectors of the optimal pairwise comparison question as determined from Section \ref{sec:question-selection} as well as the NL descriptions of each feature. The user must use these descriptions to interpret the vector comparison query, then select either `Option A' or `Option B' based on their preference.

\subsection{Human Experiment Details}

We recruited native-English-speaking human participants from Prolific \cite{Palan2017ProlificacASP}.
$\sim$40 individuals were allocated to each setting.\footnote{All human subject experiments were approved by the IRB. Additional details regarding the user studies are available in \Cref{app:user-study}}
Mirroring \citet{li2023eliciting}, we conducted our prolific experiments in two steps: \textit{(1) elicitation}: participants were asked to answer questions about their preferences for 5 minutes; \textit{(2) prediction}: participants were then presented with the 15 pairwise comparison questions and asked which of the two options they preferred. We recruited new participants for each test across all experiments. 

\subsection{Domain: Content Recommendation}

We consider the task of content recommendation--recommending news articles for a user to read. Each approach is evaluated based on its ability to predict which news articles the user would prefer on a set of 15 pairwise comparisons. Each comparison contains the lede and headline from New York Times articles, hand-collected by the authors. We select this task (1) because people's preferences have a large amount of variance and (2) to perform a direct comparison to \citet{li2023eliciting}'s methodology. Further information about the articles is available in \Cref{app:eval}.

\subsection{Evaluation}
\label{sec:eval}
We evaluate \ourmethod and the above baselines after each user interaction on a test set of 15 pairwise comparison questions, each comparing two different real-world news articles.

\paragraph{Methods}

We evaluate using two different prediction methods: in \textbf{\ourmethod prediction}, we use the \ourmethod prediction method described in~\Cref{sec:methods-prediction} to answer each test set question. In \textbf{LM prediction}, we prompt a LM to answer the test question, conditioned on the conversation history (up to some turn), similar to the evaluation approach taken by
\citet{li2023eliciting}.

\paragraph{Metrics}
After each conversation turn, we evaluate the test-set accuracy at that point in time, using one of the two methods above.
After the conversation has ended, we calculate the \textbf{Time-Integrated Delta Accuracy (TIDA)}, or the integral of the ``$\Delta$accuracy over time curve'', where $\Delta$accuracy is defined as the difference between test-set accuracy at the \textit{current} interaction step and the initial interaction step (\textit{before} any user responses).
This metric rewards eliciting user preferences quickly---even if two elicitation transcripts result in same $\Delta$ accuracy, the one that arrives at a higher $\Delta$ accuracy earlier in time is rewarded.

\section{Results}
\label{sec:results}
\begin{figure}
    \centering
    \includegraphics[width=0.99\linewidth]{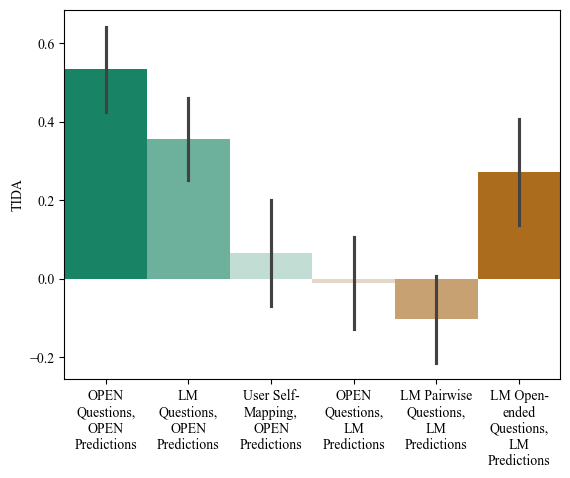}
    \caption{\textbf{Time-Integrated Delta Accuracy (TIDA) for each method}. We report the integral of the delta accuracy over time. %
    \ourmethod improves over naive prompting-based approaches, all while improving transparency and reducing computational cost. Error bars are one standard error.}
    \label{fig:content_recommendation_tida}
\end{figure}

We present the TIDA scores for \ourmethod{} against five different combinations of question generation and prediction in ~\Cref{fig:content_recommendation_tida} (refer to Sections~\ref{sec:framework} and~\ref{sec:baselines} for question generation methods and Section~\ref{sec:eval} for prediction methods). Below, we discuss the main takeaways from our results.

\paragraph{\ourmethod is better at \textit{making predictions aligned with} human preferences when compared to LM-only approaches.}
Comparing the \textit{\ourmethod Questions, \ourmethod Predictions} bar to \textit{\ourmethod Questions, LM Predictions} bar, we find that \ourmethod can make predictions that better align with human preferences than a prompted LM, from the same sequence of questions.
This implies that LMs are still subpar at \textit{in-context-learning} of human preferences from demonstrations, compared to Bayesian methods which explicitly keep track of the human's preference function; it underscores the importance of explicitly tracking human preferences.
\todo{make clear the setup differences between the pairwise -- no example articles included in conversation history, no example articles generated for LM pairwise questions, lm predictions (i.e. is directly comparable to GATE)}

\paragraph{\ourmethod is better at \textit{eliciting} human preferences compared to LM-only approaches.} Comparing the \textit{\ourmethod Questions, \ourmethod Predictions} bar to the comparable \textit{LM Pairwise Questions, \ourmethod Predictions} bar, we find that \textit{\ourmethod Pairwise Questions} outperforms \textit{LM-only Pairwise Questions} for elicitation, indicating that when asking questions, there is value in tracking human preferences and querying based on explicit notions of uncertainty.

Of course, one benefit of LM elicitation approaches is that LMs are not constrained to asking only a certain type of question. Thus, as a top-line, we elicit open-ended questions from LMs (\textit{LM Open-ended Questions, LM Predictions}). Though open-ended questions far outperform pairwise questions (both ones selected by \ourmethod and selected by the LM) when evaluated with a LM, this improvement is overshadowed by the benefit of being able to use \ourmethod \textit{prediction}, which is incompatible with LM-only elicitation methods.
However, this suggests that future lines of work could explore eliciting open-ended questions compatible with \ourmethod predictions.

\paragraph{\ourmethod is better at eliciting human preferences compared to BOED-only approaches.}
When we remove the verbalization step (\textit{User Self-mapping, \ourmethod Predictions)}, we find that users become significantly worse at accurately specifying their preferences. Thus, having a natural interface for interacting with the user, which we accomplish by verbalizing questions with a LM, is essential for being able to accurately and efficiently elicit user preferences.

\section{Analysis}
\label{sec:analysis}
\begin{figure*}
    \centering
    \includegraphics[width=\linewidth]{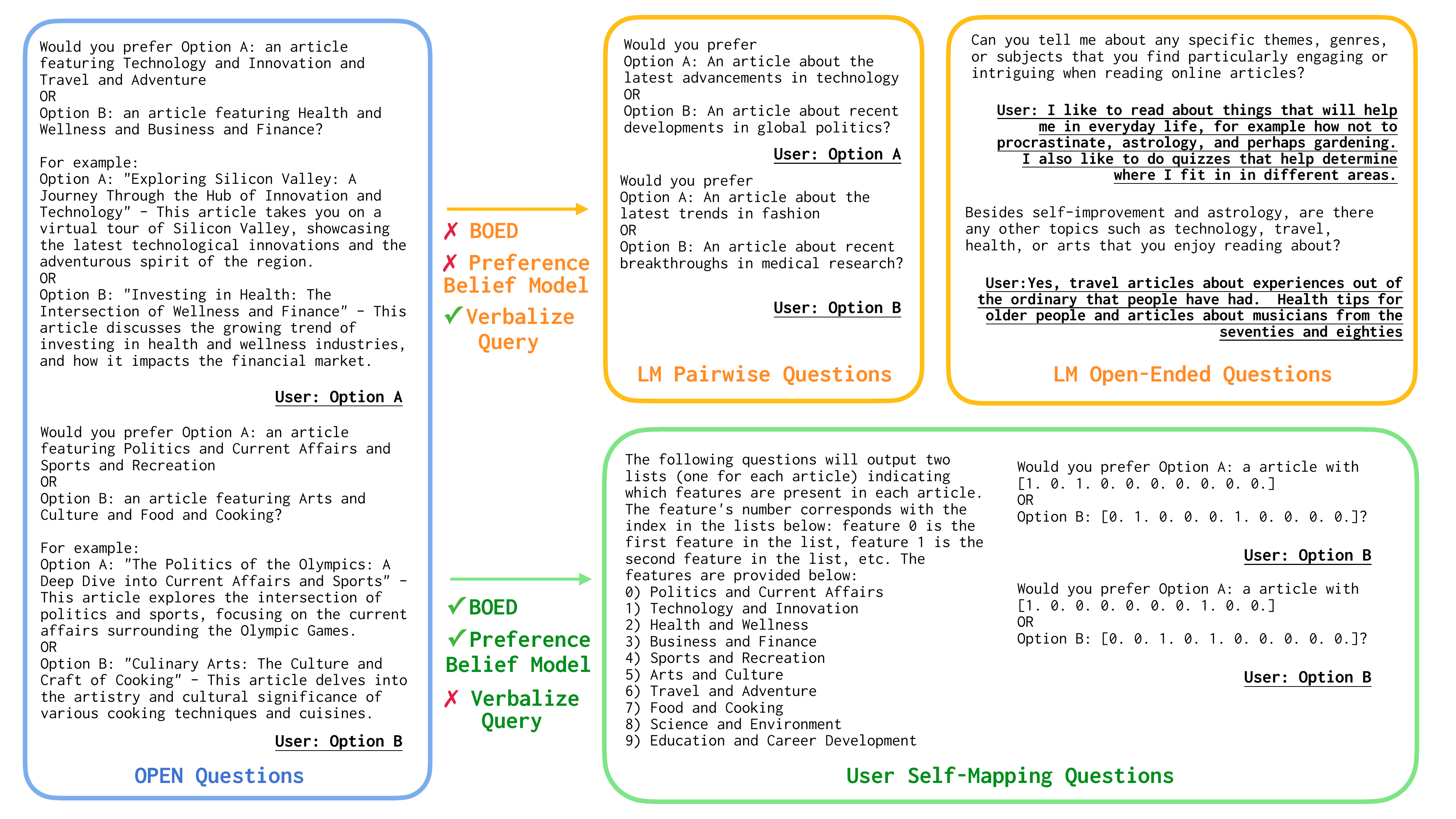}
    \caption{\textbf{Sample transcripts from \ourmethod vs. Baselines}}
    \label{fig:transcripts}
\end{figure*}

\paragraph{The Importance of Feature Weightings}
To understand the importance of \textit{feature weightings} as opposed to absolute rankings of features, we asked users to order the set of NL features by importance. We tested if a linear or exponential weighting of their self-reported rankings were able to accurately predict their preferences on the test cases. We compared the user's absolute ranking to (1) the absolute rankings of \ourmethod by averaging \ourmethod's rankings across all personas at the last interaction step and (2) the performance of \ourmethod at the last interaction step with the learned feature weights. 

\begin{figure}[ht]
    \centering
    \includegraphics[width=0.99\linewidth]{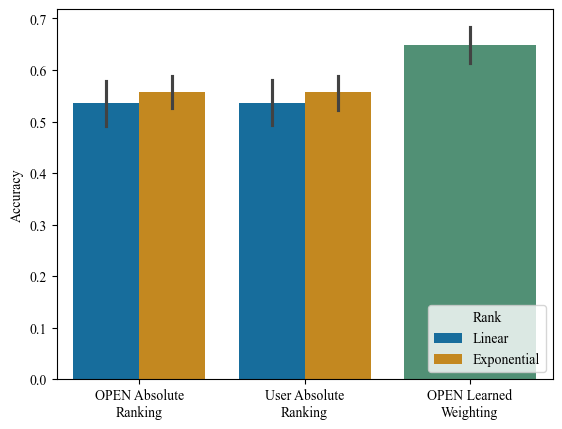}
    \caption{Accuracy for \ourmethod's absolute feature rankings, users' self reported absolute feature rankings, and \ourmethod's learned weights over features as described in \Cref{sec:analysis}. Our analysis indicates removing access to 
    precise weights hurts performance. Error bars are one standard error.}
    \label{fig:feature_ranking}
\end{figure}

Our findings, in \Cref{fig:feature_ranking}, indicate that the relative weighting of features is critical to \ourmethod's prediction, and that user's self-reported rankings are not a good indicator of their actual preferences. 
This suggests that methods that rely solely on human-written specifications, such as prompts, may not be as useful as methods that are able to keep track of the precise relative weights between features, as feature weights are typically much harder for users to specify than feature rankings.

\paragraph{Qualitative Analysis}
At the end of the study, we collected open-ended feedback from the users regarding their experience interacting with the system. Across the different elicitation methods, we found:
\kh[]{Not sure if this was the best way to convey the qualitative analysis. or if we should just remove the quotes and summarize instead}
\bzl{we can keep for now and move to appendix later}
\begin{enumerate}
    \item \textbf{LM's open-ended questions to be repetitive and overreliant on the LM's prior over user preferences}. For example one user noted: \textit{"the chat bot did not seem to pick up on many parts of my input, as though it had its own 'agenda'."} Another user observed that the model repeated questions: \textit{"the chatbot continued to ask the same exact question in the same way, it should have different variations and questions to dive deeper rather than repeating itself."} The transcripts from these users corroborate their observations. 
    \item \textbf{LM's pairwise questions (without an explicit feature set) failed to probe for \textit{feature weighting}}. A user commented that: \textit{"I’m not sure what the purpose here is, but if it was to pinpoint my interests than they should have been comparing my previous choices together as well!"}
    \item \textbf{Users struggled to self-map features to NL questions}. One user stated: \textit{"I was thoroughly confused, however I think I figured it out and clicked accordingly."}

\end{enumerate}
We also collected empirical feedback on participants' perceived effort across the different methods. We found that they reported consistent levels of mental demand across all methods (including the \textit{LM Open-ended Questions}), indicating that \ourmethod questions were not more challenging for users than other settings. Further analysis and visualization of users' empirical feedback is included in Appendix \ref{app:user-study}. Transcripts from each our settings can be found in~\Cref{fig:transcripts}.

\section{Related Work}
\paragraph{Task Ambiguity and Underspecification}
Prior research has studied ambiguity in the context of language-guided tasks~\citep{Lake2019HumanFL,tamkin2023task}. In particular, in the context of prompt engineering~\citep{brown2020language}, writing complete and unambiguous task specifications can be difficult~\citep{li2023eliciting, ieeeguide}.
Futhermore, recent developments have encouraged fine-tuning approaches such as reinforcement learning with human feedback (RLHF)~\citep{Ziegler2019FineTuningLM,christiano_deep_2017} and direct preference optimization (DPO)~\citep{rafailov2023direct} to better align LLM behaviors with humans' (underspecified) preferences. These techniques, however, all assume access to existing human preference data. Our goal in this work is to examine how to elicit such data efficiently.

\paragraph{Classical Preference Learning Techniques}
Learning human preferences is a rich area of work that has spanned many fields over past few decades. Much work has gone into \textit{optimally} querying for user preferences.\bzl{make sure we talk about BALD and BOED above}
Aside from Bayesian Optimal Experimental Design, which we use in this paper, approaches have included conjoint analysis~\citep{stern_continuum_1990,arora_improving_2001,5fd59794-2e38-36c0-8442-19944620adf1}, polyhedral methods~\citep{doi:10.1287/mksc.1060.0257}, multi-armed bandits~\citep{pmlr-v9-lu10a}, and dueling bandits~\citep{10.1214/18-AOS1772}, etc.~\cite{vayanos2021robust}

Active learning is an area of machine learning focused on interactively querying an expert to label new data points, often selected from a pool~\citep{ALsurvey}. Uncertainty-based active learning computes the next example to show the user based on explicit notions of uncertainty~\citep{cohen_heterogeneous_1994}.

However, these approaches are often only applicable to simple domains with known feature spaces (or existing pools of examples), and may not be straightforward to derive for complex domains.

\paragraph{LM Preference Learning}
More recently, \citet{li2023eliciting,piriyakulkij2023active} introduced approaches to active elicit user preferences with language models. 
They demonstrated LMs are able to actively elicit user preferences across a wide variety of domains beyond certain classical preference learning baselines.
However, our work demonstrates that LMs are still subpar at tracking and using feature weightings to ask informative questions.

\section{Discussion \& Future Work}
We presented \ourmethod, a domain-agnostic framework for combining Bayesian Optimal Experimental Design with LMs--using BOED to guide the choice of informative questions, and the LM to extract environment-relevant features and translate abstract feature queries into real NL questions. Beyond the improvements demonstrated by our findings, our system has two additional advantages over LM-only methods: (1) \textit{we can improve transparency} through extracting explicit weights and uncertainties on NL features. In our framework, the NL features from the LM accompanied by an external uncertainty-driven model conditioned on the features provide greater transparency than an LM alone. With \ourmethod, we can understand how important each feature is to the elicitation and prediction process. (2) \textit{We can reduce computational costs}. By %
representing user preferences with a linear model,
we can avoid the expensive training or inference procedures with large language models. %
In \ourmethod, the particle filter models a user's behavior while still incorporating the high-level, contextual information encoded in LMs. Furthermore, at test time, conducting inference on this linear system is orders of magnitude cheaper and yields better results than an LM. 

\ourmethod's flexibility allows for the exploration of other, more intensive preference-learning approaches such as variational Bayesian methods (\citealp[e.g.][]{foster2019variational}) or multi-armed bandits (\citealp[e.g.][]{lindner2022interactively}). We are also excited by the potential to further incorporate aspects of real-world learning environments. For example, future work could investigate using \ourmethod to develop an adaptive feature space--as the model's uncertainty increases, the model could query for and optimize new features that might better explain patterns within the data. Additionally, recent advancements in fine-tuning models from human feedback \cite{ouyang2022training, rafailov2023direct, ethayarajh2024kto} found significant differences in data curation and model performance when changing the experimental design. \ourmethod enables future research to explore the impact of this parameter in an active preference elicitation environment. 

\section*{Limitations}
Although we provide the steps to build towards better active uncertainty-driven preference learning methods, our work has several limitations. In our study, \ourmethod was constrained to pairwise queries in a content recommendation setting. Also, we operated under a fixed feature space and with a closed-source LM. Future work could explore expanding along each of these axes: investigating other preference-learning domains, incorporating open-ended questions, expanding the feature space in conversation, and testing other LMs. Futhermore, future research should explore the impact of \ourmethod across additional real-world settings and population groups. 

\section*{Ethical Considerations}
Our work presents both ethical benefits and risks. Understanding user preferences in underspecified environments is crucial to avoiding real-world issues with AI systems such as spreading hate speech, generating illicit or copyrighted content, as well as perpetuating bias and stereotypes. Furthermore, as more systems are deployed without expert-supervision, ensuring that they can align themselves with users' values is crucial to ensuring safe and healthy interactions. However, in aligning with user preferences, it is also possible that systems may align with unwanted or dangerous tendencies. Developing guidelines for models' value-systems is crucial to ensuring the long-term success of human-AI interactions. Furthermore, working with preference data inherently presents risks as it can be used maliciously to manipulate individuals. It is necessary to ensure that preference data, when used by AI systems, is collected robustly and with supervision. 

\paragraph{User Study} We used the Prolific platform \cite{Palan2017ProlificacASP} to conduct all of our human subject experiments. We paid workers in line with Prolific guidelines and solicited initial feedback via pilot studies to make the interactive experience more enjoyable for participants. Anecdotally, many participants expressed enjoying the study:
\begin{enumerate}
    \item \textit{"This was a very interesting exercise. Thank you for the opportunity."}
    \item \textit{"it was really interesting. good job."}
    \item \textit{"The task was straightforward and I liked that there were diverse topics to choose from. It felt like I had the chance to express my preferences on various subjects"}
\end{enumerate}

\paragraph{Reproducibility Statement}
We provide thorough experimental details including the user interface and LM prompts in the Appendix. We will also release our codebase and anonymized user data on GitHub. 

\paragraph{Acknowledgements}
We would like to thank Muhammed Razzak, Andreas Kirsch, Jannik Kossen, and Michael Tamkin for useful conversations as well as Lisa Schut, Gabe Grand, and Mehul Damani for feedback on earlier drafts of the paper. This material is based upon work supported by the National Science Foundation under IIS-2238240 and IIS-2212310. BZL is supported by the NDSEG Fellowship. JA is supported by the Sloan Research Fellowship.

\bibliography{anthology,custom}

\newpage
\appendix
\label{appendix}

\section{Appendix}

In the Appendix below, we provide additional  details regarding our methodology, user studies, and LM prompts.

\section{Particle Filter Details}
\label{sec:app-particle_filter}
The detailed implementation of our particle filter algorithm is below:

For the $i$-th persona $\mathbf{p}_i$ at conversational turn $t$:
\begin{enumerate}

    \item We assume the Bradley-Terry model of human preferences \cite{BradleyTerry}, where the probability of preferring Option A over Option B is directly proportional to the exponential of the difference in outcomes between Option A and Option B. We \textbf{calculate the utility of each option} for each persona:
    \[
        U(\mathbf{o}_{\{a, b\}}, \mathbf{p}_i) = \mathbf{p}_i \cdot \mathbf{o}_{\{a, b\}},
    \]
    and \textbf{compute the preference probability} %
    the persona prefers $\mathbf{o}_a$ over $\mathbf{o}_b$:
    \[
       p(y_t| \mathbf{o}_a, \mathbf{o}_b, \mathbf{p}^i_t) = \sigma(U(\mathbf{o}_a, \mathbf{p}_i) - U(\mathbf{o}_b, \mathbf{p}_i)).
    \]
    \item The \textbf{weight of each persona is set to the posterior probability} 
    \[
        \hat{w}^i_t = p(y_t | \mathbf{o}_a, \mathbf{o}_b, \mathbf{p}^i_t),
    \] 
    We then %
    \textbf{normalize the weights} over all personas:
    \[
        w^i_t = \frac{\hat{w}^i_t}{\sum_{i=1}^{N} \hat{w}^i_t}
    \] 
    
    \item Finally, we fit a Gaussian $\mathcal{N}(\mu,\sigma)$ to the re-weighted personas, by setting its parameters to
        $$\mu = \frac{\sum_{i} w^i_t}{N}$$
        $$\sigma = \sqrt{\frac{\sum_i (w_i - \mu)^2}{N}}$$

    We \textbf{resample personas} from $\mathcal{N}(\mu,\sigma)$ via systematic resampling. This step is important for preventing weight collapse.
\end{enumerate}

\section{Content Recommendation Evaluation Details}
\label{app:eval}

For evaluation in our content recommendation domain, we collected 30 articles--the article titles and lede--from the New York Times (\url{https://www.nytimes.com/}). Article topics ranged from "What to Cook This Week" to "Barbie vs. Oppenheimer" to "Fighting to Govern Myanmar." Articles were randomly paired to create a persistent set of 15 pairwise comparison questions which were used throughout all of our evaluations.

\section{User Study}
\label{app:user-study}
\subsection{Prolific Survey}
\begin{figure*}
    \centering
    \includegraphics[width=\linewidth]{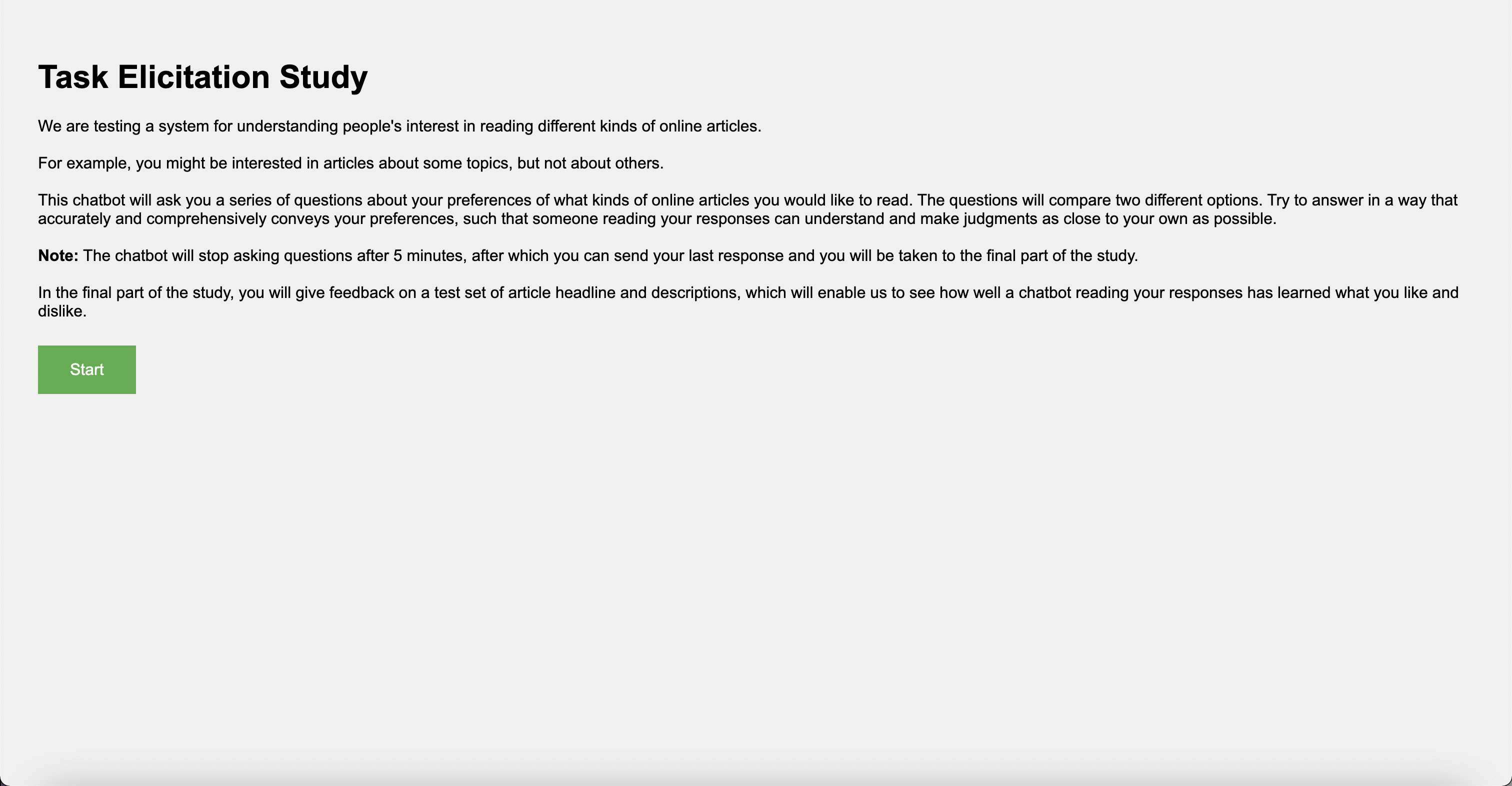}
    \caption{\textbf{Start screen}}
    \label{fig:prolific1}
\end{figure*}

\begin{figure*}
    \centering
    \includegraphics[width=\linewidth]{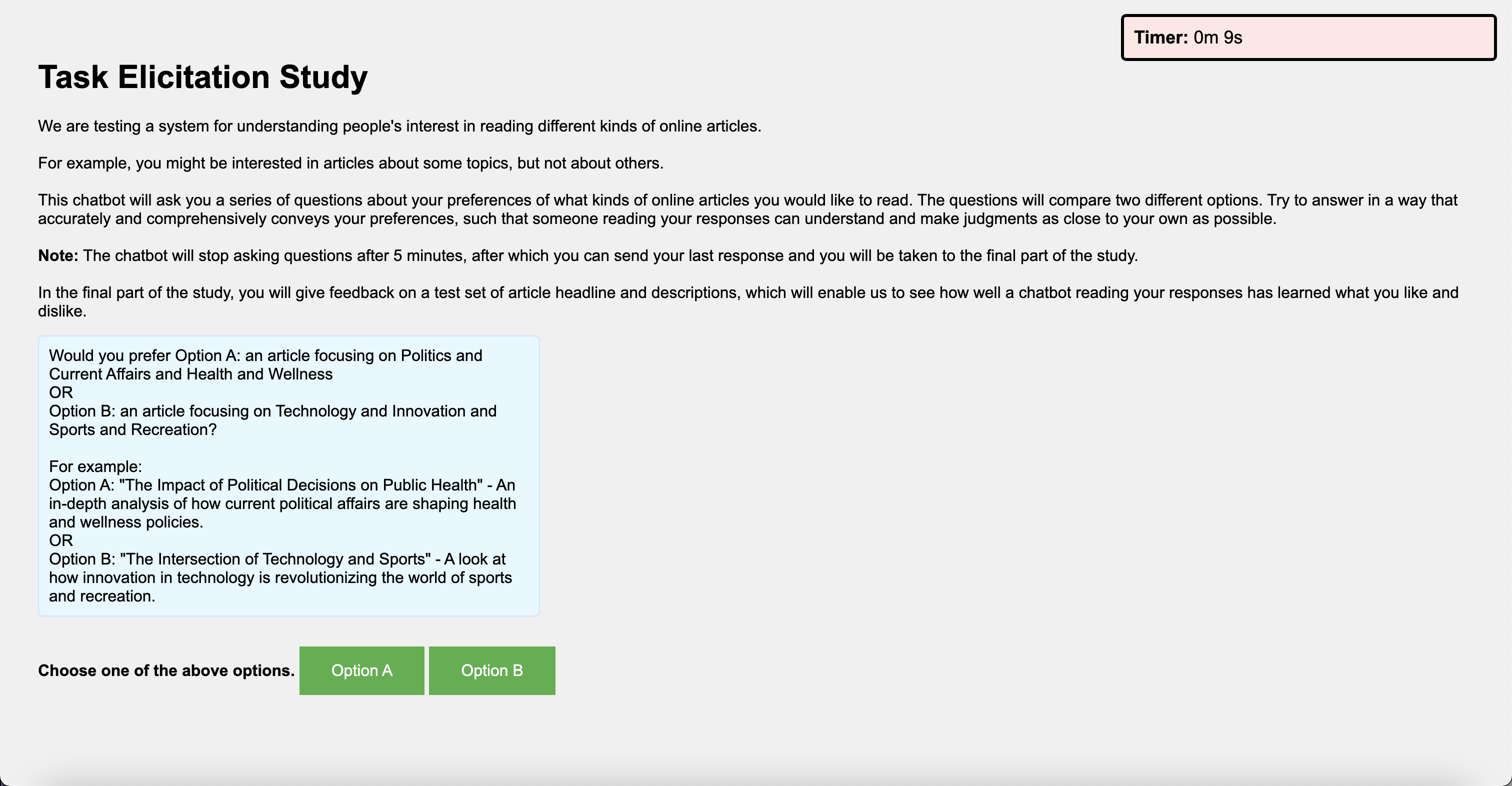}
    \caption{\textbf{Question-answering process}}
    \label{fig:prolific2}
\end{figure*}

\begin{figure*}
    \centering
    \includegraphics[width=\linewidth]{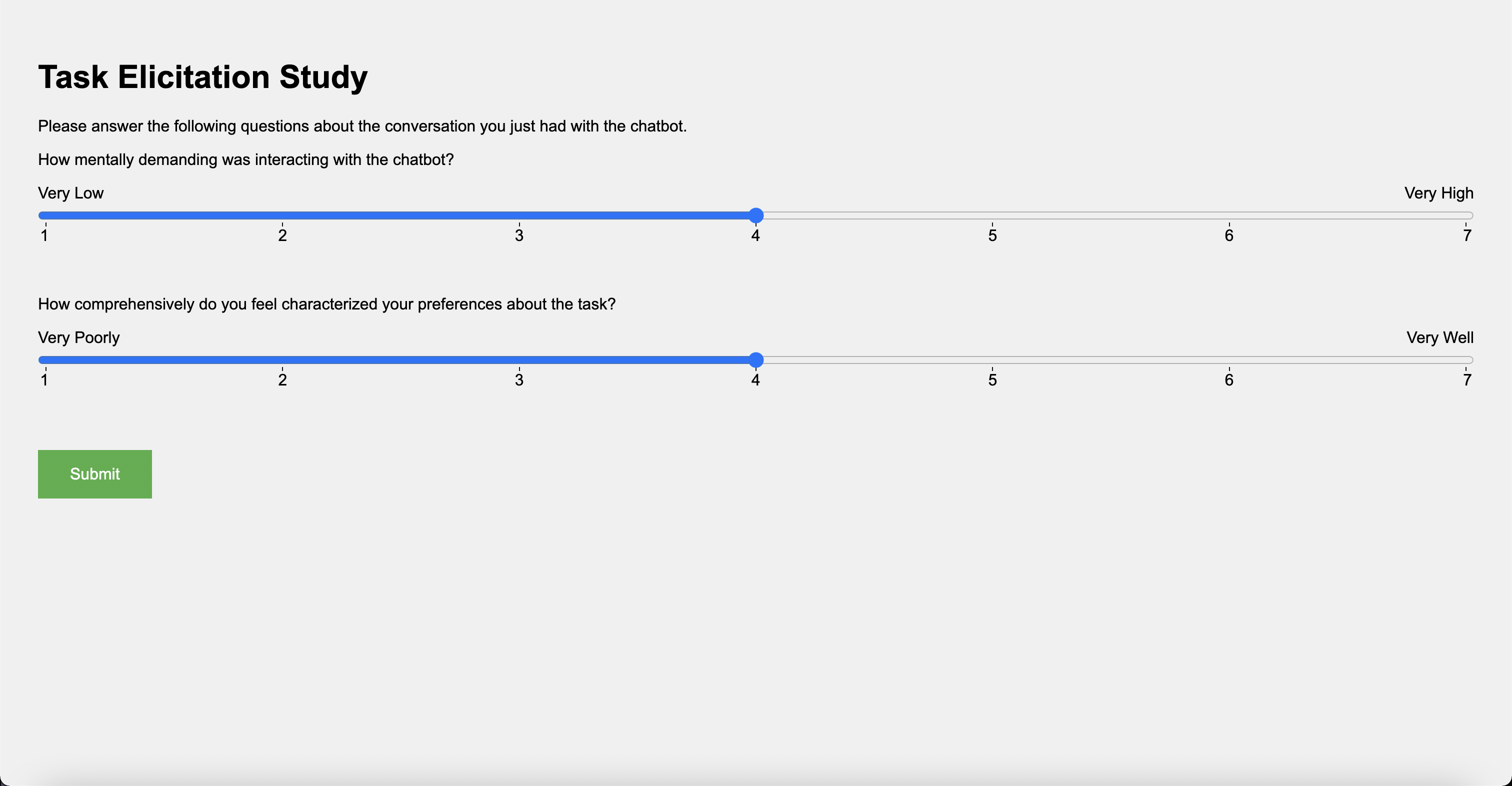}
    \caption{\textbf{Collecting user feedback metrics}}
    \label{fig:prolific3}
\end{figure*}

\begin{figure*}
    \centering
    \includegraphics[width=\linewidth]{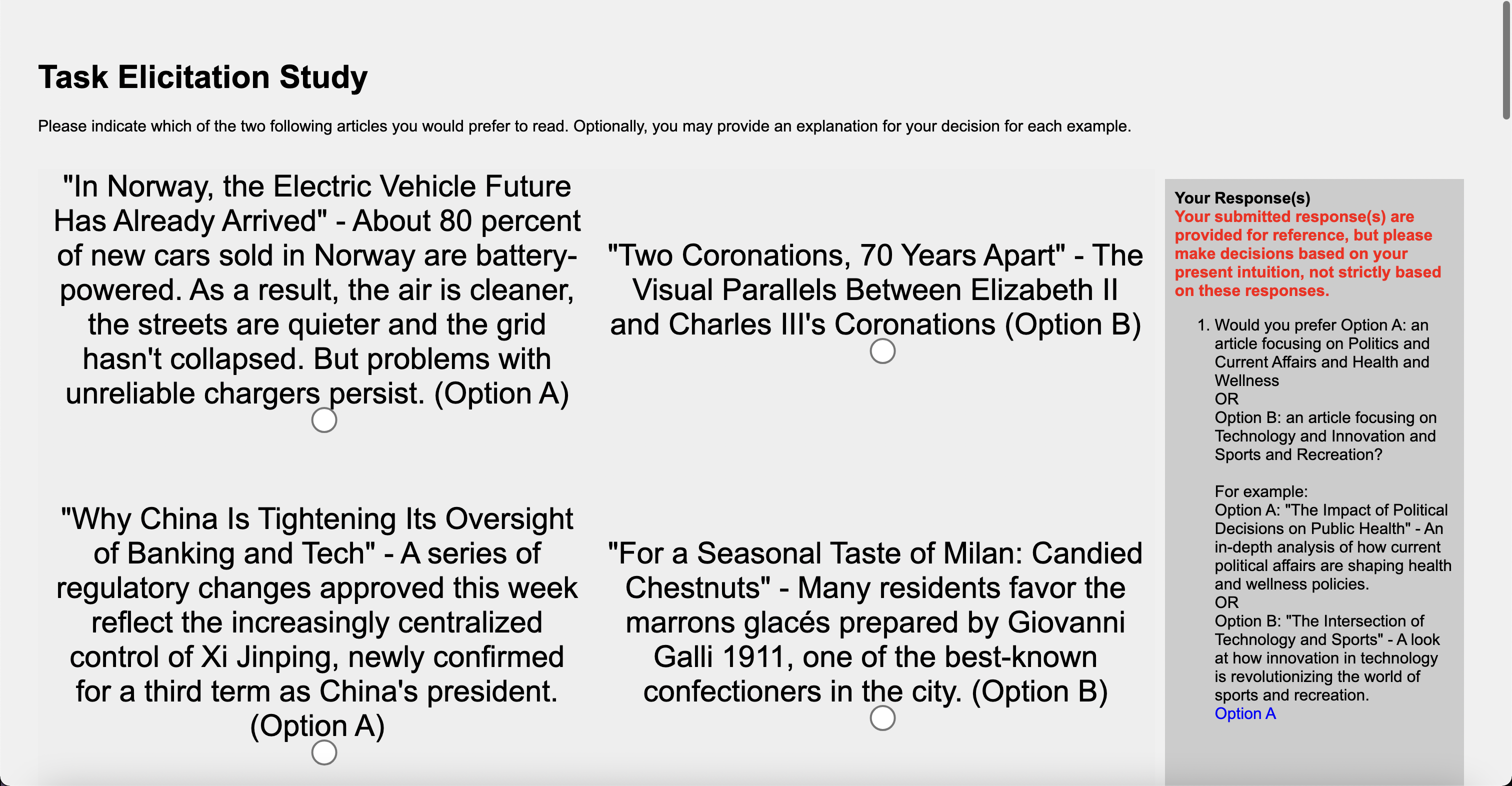}
    \caption{\textbf{Evaluation--user answering pairwise comparision test cases}}
    \label{fig:prolific4}
\end{figure*}

\begin{figure*}
    \centering
    \includegraphics[width=\linewidth]{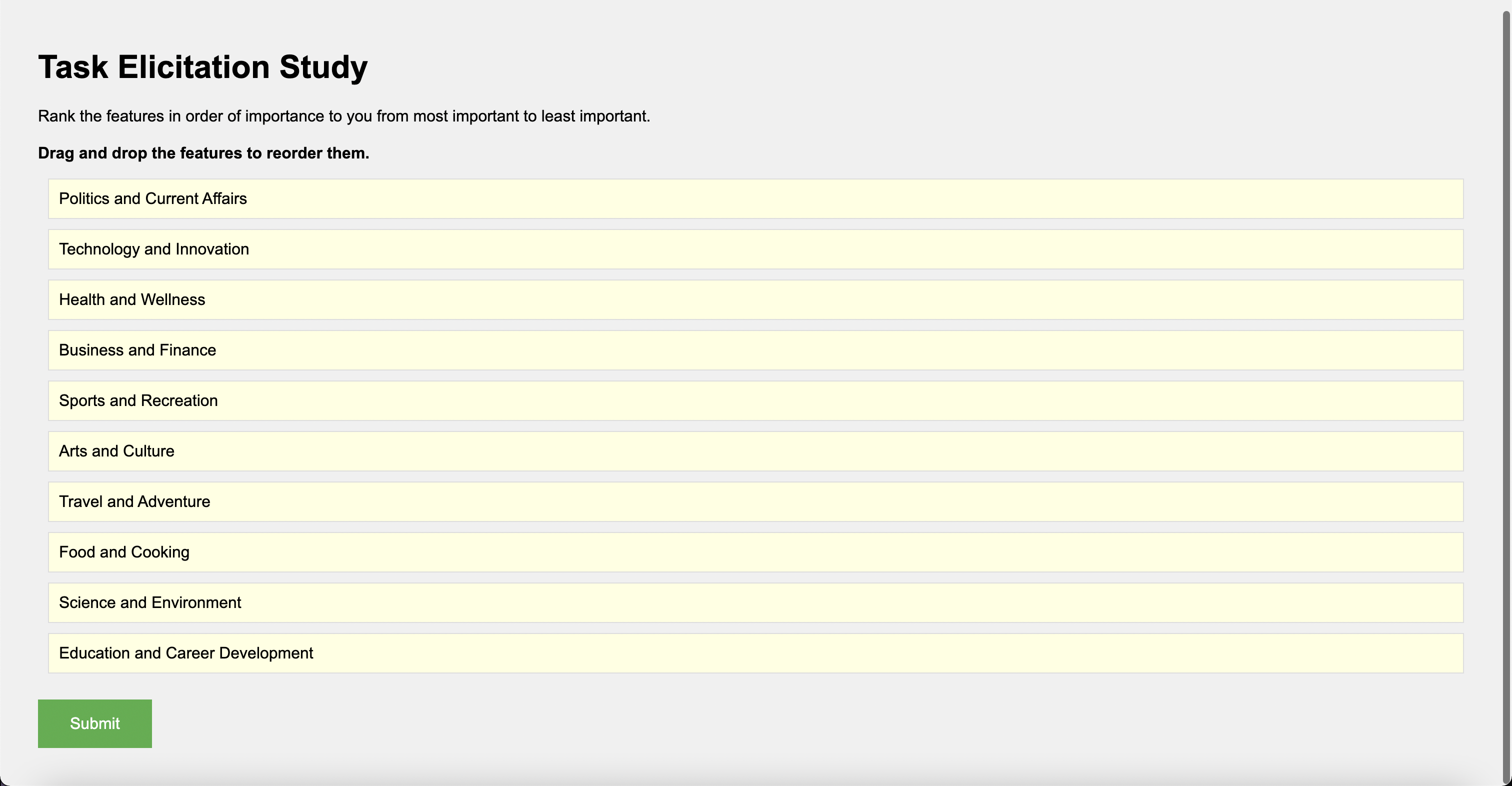}
    \caption{\textbf{User reorders NL features from most to least important}}
    \label{fig:prolific5}
\end{figure*}

Screenshots of the Prolific survey from the participants' viewpoint across different stages of the study are provided in the following pages (Figures \ref{fig:prolific1}-\ref{fig:prolific5}). We recruited English-speaking participants located in the United States. For each elicitation and prediction setting reported in \Cref{fig:content_recommendation_tida}, we recruited roughly 40 users. Specifically for each setting we had:
\begin{enumerate}
    \item \textit{OPEN Questions, OPEN Predictions}: 44 users
    \item \textit{LM Questions, OPEN Predictions}: 48 users
    \item \textit{User Self-Mapping, OPEN Predictions}: 39 users
    \item \textit{OPEN Questions, LM Predictions}: 44 users
    \item \textit{LM Pairwise Questions, LM Predictions:} 41 users
    \item \textit{LM Open-ended Questions, LM Predictions}: 39 users
\end{enumerate}

\section{Analysis: User Feedback}
\label{app:user-feedback}
\begin{figure*}
    \centering
    \includegraphics[width=\linewidth]{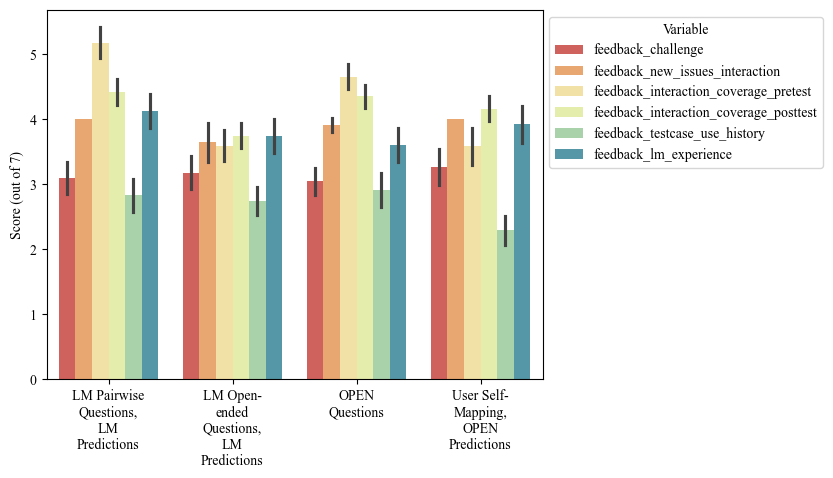}
    \caption{\textbf{User feedback analysis.} Each \textit{Variable} in the legend maps to a user prompt in \Cref{app:user-feedback}. \textit{OPEN Questions} are not more challenging for users to answer compared to other methods.}
    \label{fig:user-feedback}
\end{figure*}

\Cref{fig:user-feedback} visualizes the user feedback across multiple different axis. Each of the axis and their corresponding query for the user is below:
\begin{enumerate}
    \item feedback\_challenge: \textit{"How mentally demanding was interacting with the chatbot / writing your answer?"}
    \item feedback\_new\_issues\_interaction: \textit{"To what extent did the chatbot raise issues or aspects about your preferences that you hadn't previously considered?"}
    \item feedback\_interaction\_coverage\_pretest: \textit{"How comprehensively do you feel the chatbot's questions / your answer characterized your preferences about the task?"}
    \item feedback\_interaction\_coverage\_posttest: \textit{"After seeing the examples in the *second* part of the task, how well do you feel the chatbot / the answer you wrote (in the first part of the task) covered the important issues or aspects of these examples?"}
    \item
    feedback\_testcase\_use\_history: \textit{"When performing the *second* part of the task, to what extent did you refer back to your conversation history / answer from the first part of the task?"
    \item feedback\_lm\_experience: "How much experience have you had (if any) with interacting with language models (e.g. ChatGPT, GPT4, etc.)?"}
\end{enumerate}

As we can see, \ourmethod was not more challenging than the other methods across the different axes. Unsurprsingly, in the \textit{User-Self Mapping} case, users went back to look at their conversation history much less frequently than other in other methods. Interestingly, users felt that LM Pairwise questions covered their preferences well pre-test, but this ranking dropped significantly post-test (and did not translate to a higher accuracy on the test set).

\label{app:analysis}

\section{LM Prompts}
\label{app:prompts}

Below, we include the prompts used for the different LM-based tasks. We queried OpenAI's GPT-4 \citep{openai2023gpt4} via the API. All queries were made with temperature 0. Our prompt formatting follows previous work on verbalizing LM confidence \cite{tian2023just}.

\subsection{LM Open-ended Questions}

\begin{lstlisting}
Your task is to learn what topics a user is interested in reading online articles about. People's interests are broad, so you should seek to understand their interests across many topics; in other words, go for breadth rather than depth. Do not assume a user has given a complete answer to any question, so make sure to keep probing different types of interests.

Previous questions: {interaction_history_formatted}.

Generate the most informative open-ended question that, when answered, will reveal the most about the desired behavior beyond what has already been queried for above. Make sure your question addresses different aspects of their preferences than the questions that have already been asked. At the same time however, the question should be bite-sized, and not ask for too much at once. Phrase your question in a way that is understandable to non-expert humans; do not use any jargon without explanation. Generate the question and nothing else.

Provide your output in the format:

Question: <question>
\end{lstlisting}

\subsection{LM Pairwise Questions, LM Predictions}

\begin{lstlisting}
Your task is to learn what topics a user is interested in reading online articles about. People's interests are broad, so you should seek to understand their interests across many topics; in other words, go for breadth rather than depth. Do not assume a user has given a complete answer to any question, so make sure to keep probing different types of interests.

Previous questions: {interaction_history_formatted}.

Generate the most informative pairwise comparison question that, when answered, will reveal the most about the desired behavior beyond what has already been queried for above. Make sure your question addresses different aspects of their preferences than the questions that have already been asked. At the same time however, the question should be bite-sized, and not ask for too much at once. Phrase your question in a way that is understandable to non-expert humans; do not use any jargon without explanation. Generate the pairwise comparison question and nothing else.

Provide your output in the format:

Would you prefer
Option A: <first article option>
OR
Option B: <second article option>?
\end{lstlisting}

\subsection{LM Pairwise Question, \ourmethod Predictions}

\begin{lstlisting}
Your task is to learn what topics a user is interested in reading online articles about. People's interests are broad, so you should seek to understand their interests across many topics; in other words, go for breadth rather than depth. Do not assume a user has given a complete answer to any question, so make sure to keep probing different types of interests.
    
Previous questions: {interaction_history_formatted}.

The following {num_features} features describe people's preferences for different kinds of articles: {features}.

Generate the most informative pairwise comparison question that, when answered, will reveal the most about the desired behavior beyond what has already been queried for above.

Make sure your question addresses different aspects of their preferences than the questions that have already been asked. At the same time however, the question should be bite-sized, and not ask for too much at once. Phrase your question in a way that is understandable to non-expert humans; do not use any jargon without explanation.

Then create two specific, real-world examples of news articles matching the description you provided. For each feature you use, make sure you match the wording of the feature verbatim to the list above. Provide your output in the format:

Would you prefer Option A: an article with [features for option A]
OR
Option B: an article with [features for option B]?

For example: [Option A title and simple, one sentence description]
OR
Option B: [Option B title and simple, one sentence description]
\end{lstlisting}

\subsection{Extracting Environment Features}
\begin{lstlisting}
Your task is to learn what topics a user is interested in reading online articles about. Enumerate 10 binary topics (features) that may impact user's decisions when choosing which article to read. People's interests are broad, so you should seek to understand their interests across many topics; in other words, go for breadth rather than depth.

Only include the most important features in your list. Do not include features for which the user's preference would be obvious. Order the features from the most to least likely of interest.

Your output should be in the following format:
1) <first feature>
2) <second feature>
\end{lstlisting}

\subsection{Mapping BOED Pairwise Comparision to NL}
\begin{lstlisting}
Create two specific, real-world examples of news articles someone might be interested in reading based on the following question which juxtaposes two different news articles:
"Would you prefer Option A: an article with {pairwise_comparison_0}
OR
Option B: {pairwise_comparison_1}?".
This question instantiates two articles based on their feature values. Features lie on a spectrum ranging from 0.0 to 1.0, where 0.0 corresponds to the absence of that feature and 1.0 indicates an extremely high presence of it.

Make sure to maintain the relative difference between the two articles when generating the descriptions.

Provide your output in the format:
Option A: [Option A title and simple, one sentence description]
OR
Option B: [Option B title and simple, one sentence description]
\end{lstlisting}

\subsection{LM Evaluation}

\begin{lstlisting}
Provide your best guess and the probability that it is correct (0.0 to 1.0) for the following question. Give ONLY the guess and probability, no other words or explanation. If you are unsure take your best guess (between Option A and Option B). For example:

Guess: <most likely guess--either Option A or Option B--as short as possible; not a complete sentence!>
Probability: <the probability between 0.0 and 1.0 that your guess is correct, without any extra commentary whatsoever; just the probability!>.

A user has a particular set of preferences over what articles they would like to read. They have specified their preferences in a conversation below:
{preferences}

The question is: Based on these preferences, which of the following two articles would the user prefer?

Option A: {test_case_0}
OR
Option B: {test_case_1}
\end{lstlisting}

\end{document}